\documentclass{article}

\usepackage[nonatbib,preprint]{neurips_2021}

\usepackage[utf8]{inputenc} 
\usepackage[T1]{fontenc}    
\usepackage[pagebackref=true,breaklinks=true,colorlinks,bookmarks=false]{hyperref}       
\usepackage{url}            
\usepackage{booktabs} 
\usepackage{amsfonts}       
\usepackage{nicefrac}       
\usepackage{microtype}      
\usepackage{xcolor}         
\usepackage{enumitem}
\usepackage{subcaption}
\usepackage{graphicx}
\usepackage{multirow}
\usepackage{bm}
\usepackage{amsmath}
\usepackage{dsfont}
\usepackage{amsthm}

\newcommand{\E}{\mathbb{E}}
\newcommand{\D}{\mathcal{D}}
\newcommand{\logit}{\mathbf{logit}}
\newcommand{\1}{\mathds{1}}

\newtheorem{theorem}{Theorem}
\newtheorem{hypothesis}{Hypothesis}
\newtheorem{corollary}{Corollary}
\newtheorem{conjecture}{Conjecture}

\title{Improving Multi-Modal Learning\\ with Uni-Modal Teachers}
\author{%
  Chenzhuang Du$^{1,*}$, Tingle Li$^{1,4,*}$, Yichen Liu$^{1,*}$, Zixin Wen$^{2}$, \\
  \textbf{Tianyu Hua$^{1,4}$, Yue Wang$^3$, Hang Zhao$^{1,4}$} \\
  \small{$^1$IIIS, Tsinghua University \quad $^2$UIBE Beijing}\\
  \small{$^3$Massachusetts Institute of Technology \quad $^4$Shanghai Qi Zhi Institute}\\
  \small{\texttt{\{ducz20@mails,hangzhao@mail\}.tsinghua.edu.cn}}
}

\begin{document}

\renewcommand{\thefootnote}{\fnsymbol{footnote}}
\footnotetext[1]{Equal Contribution.}

\maketitle

\begin{abstract}

Learning multi-modal representations is an essential step towards real-world robotic applications, and various multi-modal fusion models have been developed for this purpose. However, we observe that existing models, whose objectives are mostly based on joint training, often suffer from learning inferior representations of each modality. We name this problem \textbf{Modality Failure}, and hypothesize that the imbalance of modalities and the implicit bias of common objectives in fusion method prevent encoders of each modality from sufficient feature learning. To this end, we propose a new multi-modal learning method, \textbf{Uni-Modal Teacher}, which combines the fusion objective and uni-modal distillation to tackle the modality failure problem. We show that our method not only drastically improves the representation of each modality, but also improves the overall multi-modal task performance. Our method can be effectively generalized to most multi-modal fusion approaches. We achieve more than 3\% improvement on the VGGSound audio-visual classification task, as well as improving performance on the NYU depth V2 RGB-D image segmentation task. 


\end{abstract}

\section{Introduction} \label{Sec_1: introduction}

Multi-modal signals, \textit{e.g.}, vision, sound, text, are ubiquitous in our daily life, allowing us to perceive the world through multiple sensory systems. Inspired by the crucial role that multi-modalities play in human perception and decision \cite{smith2005development}, substantial efforts have been made to build effective and reliable multi-modal systems in fields like multimedia computing \cite{aytar2016soundnet, zhao2018sound}, representation learning \cite{arandjelovic2017look, korbar2018cooperative, owens2018audio} and robotics \cite{chen2020soundspaces, gan2020look}.

Much current research on multi-modal fusion mainly revolves around the design of model architectures, such as middle fusion \cite{seichter2020efficient, wang2020deep}, late fusion \cite{wang2020makes} and attention-based fusion \cite{gan2020music, seichter2020efficient}. However, simply combining multiple modalities often results in unsatisfactory performance. \cite{wang2020makes} observed that in the video classification task, the best vision-only uni-modal network can achieve similar or even better performance than its multi-modal counterpart. They empirically examined the effects of the optimization process of different modalities and showed how the conflicts of joint multi-modal training can negatively affect the final accuracy. Such conflicts are almost inevitable when optimizing the naive joint-training objective. The inharmony of the modalities creates difficulties in multi-modal learning.

\begin{figure}[t]
	\centering
	\includegraphics[width=0.8\textwidth]{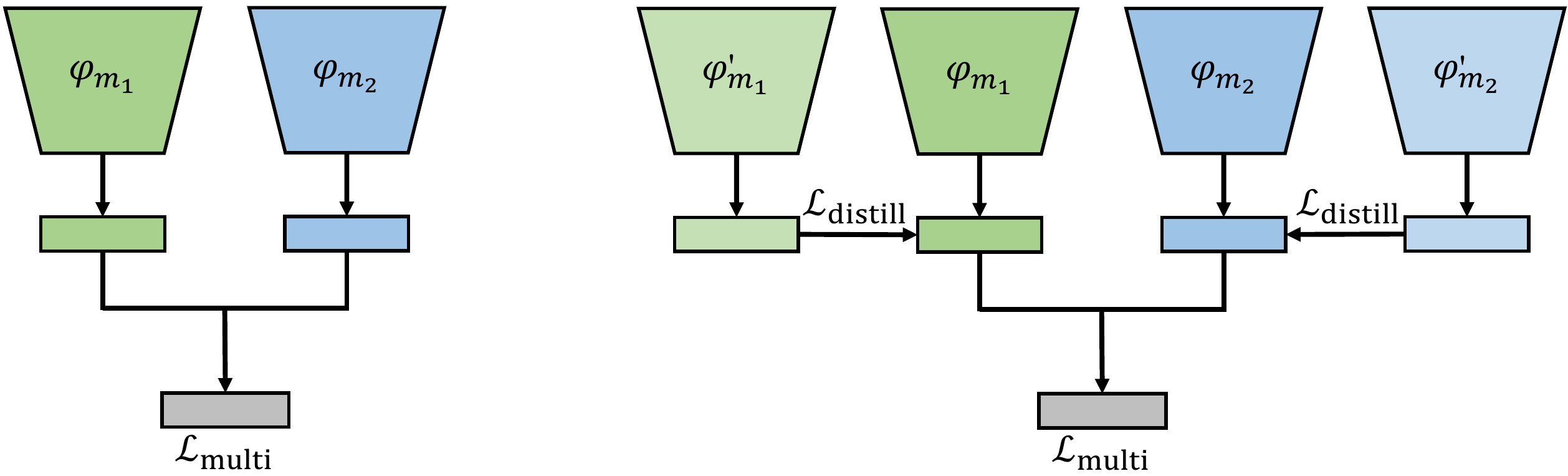}
	\caption{Model architecture of naive late fusion (left) and Uni-Modal Teacher (UMT) (right).}
	\label{fig:naive_late}
\end{figure}

In this work, we identify a significant but widely neglected weakness in previous fusion-based methods, which is termed as \textbf{modality failure}. Concretely, in fusion-based methods, the weaker modality is significantly under-trained, even when the training of the fusion model has already converged. We use the naive fusion architecture as our motivating example, which simply concatenates features extracted from all modalities (see Figure~\ref{fig:naive_late} left). As our experiments on encoder evaluation in Table~\ref{tb:modality_failure} show, even when the multi-modal classifier has already achieved 99.9\% training accuracy, the video encoder can only achieve 35\% accuracy over the training data with linear evaluation. It is worth noting that a uni-modal network trained solely on video modality can easily achieve over 99\% training accuracy.
Therefore, we hypothesize that modality failure is a main cause for the inferior performance of fusion networks.

Inspired by the recent theoretical progress of~\cite{allen2020towards}, 
we give a theoretical interpretation of the modality failure problem: the imbalance of modalities and the implicit bias of the fusion objective together lead to the insufficient feature learning of the weaker modality. In light of this interpretation, we propose \textbf{Uni-Modal Teacher (UMT)}, a distillation method whose objective is to distill the pre-trained uni-modal features to the multi-modal networks. When combined with the original fusion objective, the multi-modal networks achieve significant performance gains.
With UMT, we obtain state-of-the-art results in various multi-modal tasks, including audio-visual video classification and RGB-D semantic segmentation.


\paragraph*{Contributions.} We summarize our key contributions as follows:
\begin{itemize}[itemsep=0pt]
  \item We identify an optimization problem in multi-modal training methods called modality failure, and link it to the insufficient feature learning caused by naive joint training.
  \item To tackle this problem, we introduce a distillation method, called Uni-Modal Teacher (UMT), to alleviate modality failure and improve multi-modal performance during testing.
  \item We provide abundant analysis on modality failure and UMT, and demonstrate the effectiveness of UMT on various multi-modal tasks.
\end{itemize}


\section{Methods}

In this section, we aim to describe our findings and our proposed method UMT. First, we describe the modality failure problem of fusion-based multi-modal learning in \S\ref{Sec_2.1: observations}. In \S\ref{Sec_2.2: Uni-Modal Teachers}, we introduce Uni-Modal Teacher (UMT) which effectively solves the modality failure issue and achieves state-of-the-art performance.

\subsection{Observations of Modality Failure}
\label{Sec_2.1: observations}

\begin{table}[t!]
	\centering
	\renewcommand{\arraystretch}{1.1}
	\caption{Modality Failure: we study how well the encoders of each modality perform from various training methods. For the audio encoder, the best results are shown in bold and the worst results are underlined. ``Audio Test'' denotes the top 1 test accuracy (in \%), which is evaluated using the fixed audio encoder with a fine-tuned classifier, and similarly for the video encoders in the other column categories. Details of our experiments are in Section~\ref{Sec_4: experimental_results}.}
	\vspace{5pt}
	\begin{tabular}{c|c c c c}
		\toprule
		\multirow{2}*{\textbf{Method}}& \multicolumn{4}{c}{\textbf{Evaluation Method}} \\ \cline{2-5} & Audio Test & Video Test & Audio Training & Video Training \\
		\midrule
		Naive Fusion & 43.48 & 15.86 & 88.58 & 34.69 \\
		Audio-only Training & 46  & / & \textbf{99.95} & / \\
		Video-only Training & /  & 23.78 & / & \textbf{99.47}\\
		Gradient-Blending~\cite{wang2020makes} & 44.42 & 16.85 & 95.03 & 36.9 \\
		Audio-only Distillation & 46.28 & \underline{8.96} & 95.72 & \underline{16.45} \\
		Video-only Distillation & \underline{40.76} & 24.5 & \underline{70.95} & 82.7 \\
		\midrule
		Uni-Modal Teacher & \textbf{46.66} & \textbf{24.84} & 86.21 & 61.8 \\
		\bottomrule
	\end{tabular}
	\label{tb:modality_failure}
\end{table}

The typical approach in multi-modal learning is to combine features from encoders of different modalities to tackle a given task. Such methods are called \textbf{fusion-based methods}. Usually, the encoder networks are simultaneously updated by the gradients obtained from the training objective, \textit{i.e.}, the loss function~\cite{owens2016ambient, cartas2019does}. However, as we shall describe below, the contributions to the task from different modalities are uneven, which leads to the failure of individual modalities.

\begin{figure}[t]
    \centering
    \includegraphics[width=\textwidth]{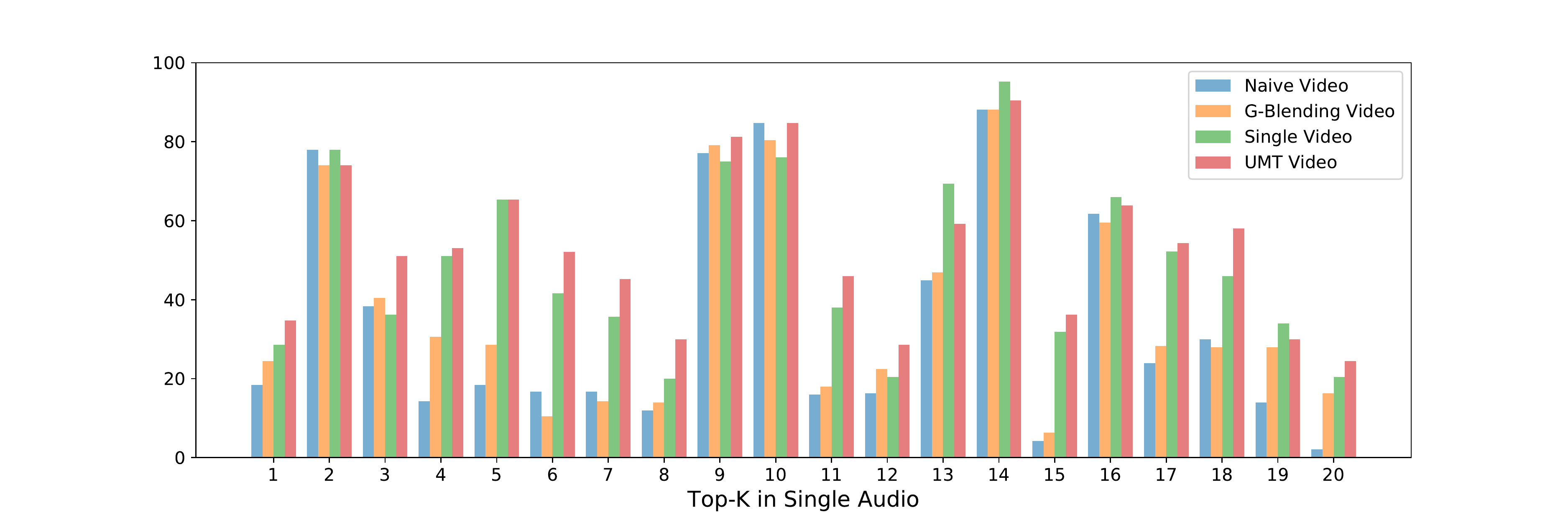}
    \caption{Video modality failure in multi-modal fusion training. We first select the top 20 test accuracy classes from uni-audio training, then evaluate different video encoders on these classes. It can be seen that the video encoder in naive fusion setting is worse than that in uni-video setting over about 16 classes, indicating that \textbf{modality failure occurs in naive fusion training}. This problem also happens in Gradient-Blending \cite{wang2020makes}, but is considerably alleviated by our UMT method.}
    \label{fig:weak_video}
\end{figure}

\paragraph{What is modality failure in fusion-based methods.} We shall illustrate that fusion-based methods cannot fully exploit the potentials of multi-modal learning, especially the \textbf{weaker modality} which contributes less to the task in terms of testing accuracy.  We call it \textbf{modality failure}, where the weaker modality, in our case the video modality, is significantly under-trained, even when the training of the fusion model has already converged. Experimental evidences are presented below:
\begin{itemize}
    \item As shown in Figure~\ref{fig:weak_video}, among the classes in which the audio network trained over uni-modal data can achieve good accuracy, the video network trained by the naive fusion method falls behind its uni-modal counterpart. Specifically, the mean accuracy on these classes in Figure~\ref{fig:weak_video} of four types of video encoder are 33.79\%, 36.92\%, 49.05\%, and 53.13\% respectively, where the gap between naive fusion video and uni-video is 15.26\%.
    \item Moreover, as Table~\ref{tb:modality_failure} shows, the naively trained video encoder achieves less than 35\% accuracy over the training data, and less than 16\% accuracy over the testing data, which are significantly lower than the performance of the video encoder trained over uni-modal data.
\end{itemize}
In order to solve the modality failure problem, we propose our Uni-Modal Teacher method.

\subsection{Uni-Modal Teacher}
\label{Sec_2.2: Uni-Modal Teachers}

Before describing our UMT method, we formally define the naive fusion approach as a base method, which we present below.

\paragraph{Naive fusion methods.} As shown in Figure~\ref{fig:naive_late} (left), naive fusion can be described as follows:\footnote{Without loss of generality, we describe classification with two modalities for the simplicity of exposition.} for the $k$-classification problem, given a training set $\mathcal{Z} = \{X_i,y_i\}_{i\in[n]} $, where the inputs $X_i = (X_i^{m_1},X_i^{m_2})$ are composed of two modalities $m_1$ and $m_2$ and $y_i \in [k]$, we use two neural network encoders $\varphi_1(W_1,\cdot) $ and $\varphi_2(W_2,\cdot)$ to map the inputs to $ X_i^{m_1} \xrightarrow{\varphi_1} \varphi_1(W_1,X_i^{m_1}) \in\mathbb{R}^{d_{\varphi_1}} $ and $X_i^{m_2} \xrightarrow{\varphi_2} \varphi_2(W_2,X_i^{m_2}) \in\mathbb{R}^{d_{\varphi_2}}$. Here $W_1, W_2$ are the weights of $\varphi_1$ and $\varphi_2$ respectively. Now by denoting the final linear classifier as $ \theta = (\theta_1,\dots,\theta_k)^{\top} \in \mathbb{R}^{k \times (d_{\varphi_1} + d_{\varphi_2})} $, we aim to solve the following optimization problem using SGD (or other variants of the gradient method):
\begin{align}\label{def-eq:naive-fusion-objective}
    \min_{\theta,W_1,W_2} L(\theta,W_1,W_2) & = \frac{1}{n}\sum_{i\in[n]} - \log \frac{e^{F_y(X)}}{\sum_{j\in[k]}e^{F_j(X)}}
\end{align}
where the function $F_y(X) := \langle \theta_{y}, (\varphi_1(W_1,X_i^{m_1}), \varphi_2(W_2,X_i^{m_2})) \rangle$. Nevertheless, as we have shown in \S\ref{Sec_2.1: observations}, such naive fusion methods will result in modality failure.


\paragraph{Uni-Modal Teacher (UMT)} The essential idea here is to distill features from well-trained uni-modal encoders to under-trained multi-modal encoders. As shown in Figure~\ref{fig:naive_late} (right), UMT involves an initial stage of uni-modal pre-training followed by a stage of distillation and multi-modal fusion. In UMT, we assume our teacher encoders $\varphi_1^{*} (\cdot),\varphi_2^{*} (\cdot)$ are pre-trained as follows: letting $s \in [2]$ indicate the modality, we solve the following uni-modal learning problem:
\begin{align}\label{def-eq:uni-modal-obj}
    \min_{\tilde{\theta}_s, \varphi_s}L(\tilde{\theta}_s,\varphi_s) = \frac{1}{n}\sum_{i\in[n]} - \log \frac{e^{F^s_y(X)}}{\sum_{j\in[k]}e^{F^s_j(X)}}
\end{align}
where $F^s_j(\cdot) :=  \langle\tilde{\theta}_{j,s},\varphi_{s} (\cdot) \rangle$ is the uni-modal learner. After obtaining the pre-trained $\varphi_s^{*}$, we distill their outputs to the randomly initialized encoders (\textit{i.e.}, $\varphi_1(W_1,\cdot) $ and $\varphi_2(W_2,\cdot)$) in the fusion model. More precisely, we use a $\ell_2$-objective $ \mathcal{L}_{\mathrm{distill}}(\varphi_s^{*}(X_i), \varphi_s(W_s,X_i)) := \|\varphi_s^{*}(X_i) - \varphi_s(W_s ,X_i)\|_2^2$ to capture the discrepancies between the features of our fusion encoders and pretrained encoders, and we add them to the final objective as follows:
\begin{align}\label{eq:umt}
    L_{\bm{\mathrm{UMT}}} := L(\theta,W_1 ,W_2 ) + \lambda\underset{(X_i,y_i)\sim \mathcal{Z} }{\E}[\sum_{s\in[2]}\mathcal{L}_{\mathrm{distill}}(\varphi_s^{*}(X_i), \varphi_s(W_s,X_i))]
\end{align}
where the fused classification loss $L(\theta,W_1 ,W_2 )$ is the same as \eqref{def-eq:naive-fusion-objective}, $\lambda$ is a hyper-parameter that weights the distillation loss. We update the parameters of the encoders via SGD (or other variants of the gradient method). As shown in Table~\ref{tb:modality_failure}, via such distillation procedure, the performance of the weaker modality can be significantly improved which results in the overall enhancement of the multi-modal evaluation. We have also compared UMT with other alternative approaches (Table~\ref{tb:av_result}) and evaluated the UMT method in different tasks (Table~\ref{tb:rgbd-result}) in Section~\ref{Sec_4: experimental_results}.

\section{Inspiration from Theory}
\label{Sec_3: inspirations from theory}

In this section, we elaborate how the multi-view structure in \cite{allen2020towards} can be analogous to the multi-modal learning tasks, and how they inspire us to solve modality failure via distillation from well-trained uni-modal encoders.

\subsection{Understanding Multi-modal Data Through Multi-view Structure}

The work in \cite{allen2020towards} explains how multi-view structure in the input can lead to insufficient feature learning. They formally define their multi-view data ($(X,y) \sim \D_m $) as follows: in a $k$-classification problem, they assume each class $y \in [k]$ has two class-indicating features\footnote{In \cite{allen2020towards}, they claim the two-features setting can be easily generalize to data with more features, but the proof would be much more involved and non-illustrative.} $v_{y,1}$ and $v_{y,2}$ (\textit{i.e.}, the class index is dependent on both of them). Let $\mu>0$ be a small constant. For about $1-\mu$ portion of the population data $X$ in class $y$, $X$ contains both features $v_{y,1}$ and $v_{y,2}$, and the rest of the population only has one feature $v_{y,i}, i\in [2]$. Over such multi-view data distribution, their theorem can be stated as follows: (the $F_y(\cdot)$ here is their neural network output, similar to ours)

\begin{theorem}[Theorem 1 in \cite{allen2020towards}, sketched]\label{thm:single-model}
    For some $k$-classification problems, over certain finite dataset sampled from the population of multi-view data, even if the training accuracy is $100\%$ (meaning all training data are correctly classified), with high probability it holds that \begin{align*}
        \underset{(X,y)\sim\D_m}{\mathbf{Pr}}(y \neq \arg \max_{y'}F_{y'}(X)) \geq 0.49\mu 
    \end{align*}
    where $\mu$ is the proportion of the single-view data, \textit{i.e.}, data with only one class-indicating feature. This means that the learner $F(\cdot)$ fails to learn one of the class-indicating features in most classes.
\end{theorem}

It's straightforward to link our multi-modality model to the multi-view structure. In our two modality settings, we present the following hypothesis below. Note that the features in our setting should be formulated as similar to those in \cite{allen2020towards}.

\begin{hypothesis}\label{hyp:multi-modality}
    Let $\mu > 0$ be some small constant. We assume for each multi-modal data sample $(X_i^{m_1}, X_i^{m_2}),y_i$ in the training dataset, with probability $1- \mu$ the class-indicating features are contained in both $X_i^{m_1}, X_i^{m_2}$, and with probability $\mu$ only one of $X_i^{m_1}, X_i^{m_2}$ contains the class-indicating feature. Moreover, we consider \textbf{imbalanced} multi-modal data, that is, for each class, the feature in single-feature data mostly (\textit{e.g.}, $1-o(1)$ fraction of such feature) comes from the stronger modality (\textit{e.g.}, the audio modality in our practice).
\end{hypothesis}

Given this hypothesis, we immediately obtain a corollary to Theorem~\ref{thm:single-model}. We shall give an informal analysis to explain what happens to the learning processes involved below.

\begin{corollary}
    Under similar learner architecture in \cite{allen2020towards} (where different classes use different encoders), if Hypothesis~\ref{hyp:multi-modality} holds, then we have the same generalization results for our multi-modal learner. That is,  for each class, the learner will inevitably fail to learn the features of one of the modalities (usually the weaker modality).
\end{corollary}

\begin{table}[t!]
	\centering
	\caption{Top-1 accuracy (in \%) under UMT and baseline methods on VGGSound dataset, where the best result is shown in bold. ``Dropout" denotes naive fusion with 0.5 dropout ratio; ``Pre-train + Fine-tune" represents first pre-training uni-modal encoders, then fine-turning a classifier over them; ``Modality Dropout" that randomly drops (with probability $1/3$) the outputs from one modality in every iteration; ``Self Distillation" means distilling a pre-trained naive fusion model to a new one.}
	\vspace{5pt}
	\begin{tabular}{c c c c}
		\toprule
		\textbf{Method} & \textbf{Top-1 Accuracy} & \textbf{Method} & \textbf{Top-1 Accuracy}\\
		\midrule
		Naive Fusion & 49.46$\pm$ 0.28 (Baseline) & Pre-train + Fine-tune & 50.81$\pm$0.33   \\
		Dropout\cite{srivastava2014dropout} & 49.83$\pm$0.12  & Modality Dropout & 51.37$\pm$0.76  \\
		Video-only Distillation & 49.55$\pm$0.06  & Audio-only Distillation & 48.84$\pm$0.27  \\
		Self Distillation & 49.86$\pm$0.66  &  Gradient-Blending~\cite{wang2020makes} & 50.39$\pm$ 0.21  \\
		\midrule
		Uni-Modal Teacher & \textbf{53.46$\pm$0.49 } & / & / \\
		\bottomrule
	\end{tabular}
	\label{tb:av_result}
\end{table}
\subsection{Why Naive Fusion Methods Lead to Modality Failure.}

Given the fusion objective, a straightforward observation is that the learning processes of two modalities are rather separated. Take modality $m_1$ for example, let $y \in [k]$ be a class index, we break $\theta_y $ into $ \theta_j = (\theta_{y,1}, \theta_{y,2}) $ so that we can rewrite the output as 
\begin{align*}
    F_y(X) = \langle \theta_{y,1}, \varphi_1(W_1,X_i^{m_1}) \rangle + \langle \theta_{y,2}, \varphi_2(W_2,X_i^{m_1}) \rangle.
\end{align*}
which is a linear combination of two predictors from different modalities. So the update rule of our parameters can be written as follows: denoting $\logit_{y}(F,X) := \frac{e^{F_{y}(X)}}{\sum_{j\in[k]}e^{F_{j}(X)}} $, at each iteration $t \geq 0$, with learning rate $\eta > 0$, our parameters are updated as (using GD)
\begin{align*}
    \theta_{y,1}^{(t+1)} &=  \theta_{y,1}^{(t)} - \eta  \underset{(X_i,y_i)\sim \mathcal{Z} }{\E} \left[(\1_{y_i=y} - \logit_{y}(F,X_i) )\cdot \varphi_1(W_1^{(t)},X_i^{m_1}) \right], \\
    W_{1}^{(t+1)} &=  W_{1}^{(t)} - \eta  \underset{(X_i,y_i)\sim \mathcal{Z} }{\E} \left[(\1_{y_i=y} - \logit_{y}(F,X_i) )\cdot \nabla_{W_1}\langle\theta_{y,1}^{(t)}, \varphi_1(W_1^{(t)} ,X_i^{m_1})\rangle \right],
\end{align*}
and similarly for $\theta_{y,2}^{(t)}$ and $W_{2}^{(t)} $. We can easily see that updates of $\theta_{y,1}$ and $W_{1}$ have no correlation to those of different modality (that is $\theta_{y,2}$ and $W_{2}$) other than from $\logit_y(F,X_i)$, \textit{i.e.}, from the training loss/accuracy. More specifically, under the imbalance assumption in Hypothesis~\ref{hyp:multi-modality}, the stronger modality can achieve close to $100\%$ accuracy over the training data even when the weaker modality is under-trained. When the training of all data $\{X_i,y_i\}_{i\in[n]}$ are close to convergence, we have
\begin{align*}
    \1_{y_i=y} - \logit_{y}(F,X_i) \approx 0 \ \text{ for all } i\in[n] \quad \implies\quad \|\nabla_{\theta} L \|,\ \|\nabla_{W_1} L \|,\ \|\nabla_{W_2} L \| \approx 0.
\end{align*}
That is, the algorithm can already converge, even when the encoder of the weaker modality (say $\varphi_2$) is largely under-trained. This is indeed what happens in our experiments. As shown by Table~\ref{tb:modality_failure}, the video encoders trained by the naive fusion methods and Gradient-Blending \cite{wang2020makes} cannot achieve more than $40\%$ training accuracy in single-modal evaluation. Such under-training is due to the inherent bias of the fusion methods and the imbalance of contributions of different modalities. Nevertheless, the success of knowledge distillation in \cite{allen2020towards} inspires us to propose our UMT method.

\subsection{How Distillation Helps Multi-modal Learning.} 

We first describe how knowledge distillation helps in multi-view setting, then we discuss how they inspire the UMT method. The statement on knowledge distillation in \cite{allen2020towards} can be sketched as follows:

\begin{theorem}[Theorem 3 in \cite{allen2020towards}, informal]\label{thm:distill}
    Over multi-view data, single model can learn all the class-indicating features by matching the soft labels of ensembles of $\widetilde{\Omega}(1)$ many single-models.
\end{theorem}

What happens in Theorem~\ref{thm:distill} can be sketched as follows: the soft-labels of the ensemble contain information of the neglected features in each class, thus matching the soft labels can help the single model to learn a complete set of features. 

Yet in our setting, matching the fusion output to soft labels of well-trained uni-modal networks may not help, since both networks have already achieved $100\%$ training accuracy. We instead sought a different approach: in UMT, we distill the features (\textit{i.e.}, the inputs of the last layer) from well-trained uni-modal networks to the corresponding encoders in naive fusion methods. We conjecture the following statement that inspired our method:

\begin{conjecture}
    By carefully picking the $\lambda$ parameter, the encoders trained by the UMT method can at least learn all the features learned by the corresponding teacher encoders. Furthermore, we also conjecture that UMT can help the encoders to extract features that can only be learned by multi-modal training, outperforming the pre-trained uni-modal encoders in linear evaluations.
\end{conjecture}

Indeed, by setting $\lambda \rightarrow \infty$, we can obtain a trivial solution, that is $\varphi_m = \varphi_m^{*}$, for $m \in [2]$. Then the learning problem reduces to train a linear classifier over pre-trained features. In this way UMT has enlarged the feature sets to incorporate both modalities, so that it can outperform uni-modal networks. We leave the proof of this conjecture as a future direction.\footnote{If we are using similar learner architecture as in \cite{allen2020towards} (where predictor of different classes use different encoders), then it is hopeful to use the techniques in \cite{allen2020towards} to prove our conjecture, since their results are proven for one-hidden-layer networks and their soft labels are similar to the features in our setting.}

\section{Experiments}

\begin{figure}[t]
	\centering
	\includegraphics[width=0.8\textwidth]{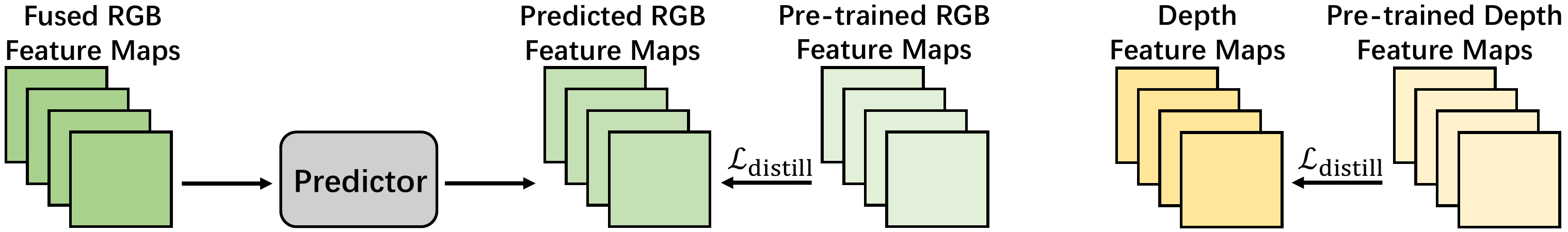}
	\caption{Model architecture of UMT for RGB (left) and depth (right) modalities.}
	\label{fig:mid}
\end{figure}

\label{Sec_4: experimental_results}
In this section, we test the effectiveness of UMT on two standard multi-modal tasks: audio-visual classification and RGB-Depth semantic segmentation. In ~\S\ref{sec:audio_visual}, we verify our method on audio-visual classification and report the performance. we also compare different solutions for modality failure. In \S\ref{sec:rgbd}, we demonstrate UMT in middle fusion task, \textit{i.e.}, semantic segmentation. In each sub-section, we first introduce the dataset, then describe the implementation details on how to apply our method to a specific task. We also provide further empirical demonstrations of modality failure through evaluating the performance of different encoders, as well as showing how UMT tackles this issue. Here, all the networks were built and trained using PyTorch~\cite{paszke2019pytorch} and all experiments were done by using one NVIDIA GeForce RTX 3090 GPU.

\subsection{Audio-visual Classification Experiment}\label{sec:audio_visual}

\paragraph{Dataset.} VGGSound dataset \cite{chen2020vggsound}, which contains over 200k video clips for 309 different sound classes, is used for evaluating our method. It is an audio-visual dataset \textit{in the wild} where each object that emits sound is also visible in the corresponding video clip, making it suitable for scene classification tasks. Please note that some clips in the dataset are no longer available on YouTube, and we actually use about 175k videos for training and 15k for testing, but the number of sound classes remains the same. 

\paragraph{Implementation details.} Two ResNet18~\cite{he2016deep} backbones are employed as our video and audio encoders respectively (whether 3D CNN is needed depends on the input), aiming at extracting visual and acoustic features simultaneously. Then we apply the late fusion, \textit{i.e.}, fusing the visual and acoustic features before the linear classifier, to incorporate video and audio information. We design a preprocessing paradigm to improve training efficiency as follows: (1) each video is interpolated to 256$\times$256 and saved as stacked images; (2) each audio is first converted to 16 kHz and 32-bit precision in the floating-point PCM format, then randomly cropped or tiled to a fixed duration of 10s. For video input, 32 frames are uniformly sampled from each clip before feeding to the video encoder. While for the audio input, a 1024-point discrete Fourier transform is performed using nnAudio \cite{cheuk2020nnaudio}, with 64 ms frame length and 32 ms frame-shift. And we only feed the magnitude spectrogram to the audio encoder. Please note that we do not use any kinds of data augmentation for both video and audio input. Besides, all the models are trained with a batch size of 24 and an initial learning rate of 1e-3 for 20 epochs, using the Adam optimizer~\cite{kingma2014adam}. We also apply a fixed learning rate scheduler, \textit{i.e.}, decay the learning rate by 0.1 for every 5 epochs and 10 epochs on UMT and baseline methods respectively. We use MSE loss as our distillation objective and we set the $\lambda$ parameter in Equation \eqref{eq:umt} to 50. Unless otherwise stated, all distillations were performed on feature-level.  



\paragraph{Baseline methods.} 
We compare UMT with other baseline methods, \textit{e.g.}, naive fusion, Gradient-Blending \cite{wang2020makes}, fine-tuning a multi-modal classifier from encoders pre-trained on uni-modal data, dropout \cite{srivastava2014dropout}, self distillation \cite{zhang2019your} and modality dropout. Besides, we also implement two other methods, \textit{i.e.}, video-only distillation and audio-only distillation, to further support our hypothesis. 

\begin{table}[t]
	\centering
	\renewcommand{\arraystretch}{1.1}
	\caption{Model performance comparison under UMT and ESANet on NYU-DepthV2 RGB-Depth semantic segmentation task.}
	\vspace{5pt}
	\begin{tabular}{c c c c}
		\toprule
		\multirow{2}*{\textbf{Initialization}} & \multicolumn{2}{c}{\textbf{Training Setting}}  & \multirow{2}*{\textbf{Improve}} \\ \cline{2-3} & ESANet~\cite{seichter2020efficient} & UMT\\
		\midrule
		From Scratch & 38.59 & 40.45 & \textbf{1.86} \\
		ImageNet Pre-train  & 48.48  & 49.14 & \textbf{0.66} \\
		\bottomrule
	\end{tabular}
	\label{tb:rgbd-result}
\end{table}

\paragraph{Results.}
From the results presented in Table~\ref{tb:av_result}, it is clear that UMT outperforms all the baseline methods by a large margin, suggesting that distillation could circumvent modality failure in joint training. In addition to showing that UMT can greatly improve performance, we also compare different methods to help us understand the fundamental issues of multi-modal learning:
\begin{itemize}
    \item Fine-tuning a multi-modal classifier over pre-trained uni-modal encoders can outperform naive fusion. This implies the modality failure problem is critical in fusion-based methods.
    \item UMT outperforms finetuning over pre-trained uni-modal encoders, showing that multi-modal learning can help the encoders to learn features that are unique to corresponding multi-modal tasks, since UMT differs from uni-modal pre-training by the multi-modal objective.
    \item We compare the results of vanilla Dropout \cite{srivastava2014dropout} with Modality Dropout. Our results show that modality-wise dropout is more effective in multi-modal learning. Also, self-distillation performing worse than UMT also implies we should pay more attention to modality-wise features.
    \item Distilling from only one modality can deteriorate the performance of other modalities. The results in Table~\ref{tb:modality_failure} and our analysis in Section~\ref{Sec_3: inspirations from theory} show that modality failure is caused by the learning process of strong modalities. Thus, only distilling from one modality is not able to avoid modality failure.
\end{itemize}

\paragraph{Encoder evaluation.}
We also do another experiment, \textit{i.e.}, fix the encoder from different training methods and fine-tune a classifier, which allows us to have a better understanding of why UMT outperforms baseline methods. As we can see in  Table~\ref{tb:modality_failure}, UMT achieves the best results on both audio and video modalities, manifesting that it could keep the representation power of each modality. It is worth noting that, while performing distillation on a single modality, the representation of the corresponding modality will be maintained, but that of the other modality will be significantly degraded. Hence, we further demonstrate that the inferior performance of naive fusion is because the stronger modal encoder would rapidly learn a powerful feature to fit the samples, while the weaker encoder cannot learn enough, which leads to modality failure.

\subsection{RGB-Depth Semantic Segmentation Experiment}
\label{sec:rgbd}
\paragraph{Dataset.} We evaluate our approach on the commonly used RGB-D multi-class indoor semantic segmentation dataset, namely NYUv2 dataset, which contains 1449 indoor RGB-Depth data totally and we use 40-class label setting. The number of training set and testing set is 795 and 654 respectively. 

\paragraph{UMT in semantic segmentation.} In contrast to the late fusion classification task, the RGB-Depth semantic segmentation belongs to middle fusion. Since features generated by each layer matter, we distill multi-scale depth feature maps using the MSE loss. For feature maps from the RGB encoder, however, since they are generated by fusing RGB and depth modalities, we cannot distill RGB feature maps directly like depth feature maps. To mitigate this effect, we curate predictors, namely 2 layers CNNs, aiming to facilitate the fused feature maps to predict the RGB feature maps trained by the RGB modality before distillation. The full schematic diagram is presented in Figure~\ref{fig:mid}.
\paragraph{Implementation details.} Our segmentation UMT is based on a state-of-the-art method, \textit{i.e.}, ESANet \cite{seichter2020efficient}. For RGB modality, the ResNet34 backbone, downsampling method, and contextual module are employed following \cite{orsic2019defense, zhao2017pyramid}. When it comes to the decoder, it receives skip connections from the encoder, which is akin to U-Net \cite{ronneberger2015u}. For Depth modality, we leverage another encoder that focuses on extracting geometric information and fuse it with RGB feature maps at the five different scales using the attention mechanism. We keep all hyper-parameters (\textit{e.g.}, learning rate, optimization schemes, dropout ratio, \textit{etc}.) the same as the official implementation\footnote{\url{https://github.com/TUI-NICR/ESANet}}.

\begin{table}[t]
	\centering
	\renewcommand{\arraystretch}{1.1}
	\caption{Ablation study on RGB-Depth semantic segmentation setting. “Initialization” indicates how weights are initialized for the network, “from Depth” represents end-to-end training with a depth-only segmentation network, and “from RGB+depth” refers to freezing the depth encoder from ESANet \cite{seichter2020efficient} then fine-tuning with a new decoder.}
	\vspace{5pt}
	\begin{tabular}{c c c c}
		\toprule
		\multirow{2}*{\textbf{Initialization}} & \multicolumn{2}{c}{\textbf{Training Setting}}  & \multirow{2}*{\textbf{Drop}} \\ \cline{2-3} & from Depth & from RGB+Depth\\
		\midrule
		From Scratch & 32.69 & 28.53 & \textbf{-4.16} \\
		ImageNet Pre-train  & 39.45  & 34.73 & \textbf{-4.72} \\
		\bottomrule
	\end{tabular}
	\label{tb:depth-ablation}
\end{table}

\paragraph{Results.} It is common to use ImageNet pre-trained parameters as the initialization weights to achieve higher performance \cite{orsic2019defense}. To this end, we train the network on both scenarios: (1) training from scratch on NYUv2; (2) pre-training on ImageNet followed by fine-tuning on NYUv2. As shown in Table~\ref{tb:rgbd-result}, UMT improves the mean Intersection over Union (mIoU) metric on both settings, demonstrating its effectiveness. Furthermore, the increment discrepancy on training from scratch is more apparent than that on fine-tuning, manifesting that each encoder in multi-modal architecture could learn better general-purpose representations \cite{huh2016makes} after pre-training.


\paragraph{Encoder evaluation.} To explore whether modality failure also exists in RGB-Depth segmentation, we evaluate the depth encoder from ESANet\cite{seichter2020efficient} by freezing its parameters and fine-tuning a new decoder. We compare it with uni-depth setting. Please note that only the depth encoder is applicable for this experiment since the RGB encoder has been fused with the depth information. As shown in Table~\ref{tb:depth-ablation}, it turns out the encoder from RGB-Depth yields worse performance on both types than uni-depth setting, which verifies our hypothesis.

\section{Related Work}
\paragraph{Multi-modal fusion.} There are several different fusion methods, including middle fusion~\cite{seichter2020efficient, wang2020deep, hu2019acnet, hazirbas2016fusenet, jiang2018rednet}, late fusion~\cite{wang2020makes, atrey2010multimodal} and attention-based fusion~\cite{gan2020music, seichter2020efficient}. Specifically, ESANet~\cite{seichter2020efficient} fed depth feature maps into RGB encoder at multiple scales; CEN~\cite{wang2020deep} exchanged different modalities' channels in the middle of the encoders; Hori \textit{et al.}~\cite{hori2017attention} proposed an attention-based fusion architecture for video captioning. MFAS~\cite{perez2019mfas}, on the other hand, posed multi-modal fusion as a neural architecture search problem. More recently, Wang \textit{et al.} proposed Gradient-Blending~\cite{wang2020makes}, which leveraged concatenated features from two modalities without any aggregation in the intermediate stage, but introduced additional weighted losses for better training. 


\paragraph{Knowledge distillation.} Knowledge distillation \cite{bucilu2006model} was first introduced to compress the knowledge from an ensemble into a smaller and faster model but still preserve competitive generalization power. Hinton \textit{et al.} \cite{hinton2015distilling} proposed to use a temperature in the \textit{softmax} outputs to represent smaller probabilities, as opposed to \cite{bucilu2006model} that just matches the output logits. Romero \textit{et al.} \cite{romero2014fitnets} further extended distillation from the output labels to intermediate representation. 

\paragraph{Audio-visual learning.} 
One of the most popular multi-modal data pair is vision and audio, as they are naturally co-occurred and are recorded by video cameras simultaneously.
Researchers have explored various tasks utilizing both of them, such as representation learning \cite{arandjelovic2017look, korbar2018cooperative, owens2018audio}, scene classification \cite{chen2020vggsound, gemmeke2017audio}, vision-assisted speech recognition~\cite{afouras2018deep}, audio-visual source separation~\cite{zhao2018sound, ephrat2018looking, rouditchenko2019self}, audio source grounding~\cite{chen2021localizing, harwath2018jointly, zhao2018sound}, audio spatialization~\cite{gao20192, morgado2018self, yang2020telling}, emotion recognition~\cite{albanie2018emotion} and audio-visual navigation \cite{chen2020soundspaces, gan2020look}.

\paragraph{RGB-Depth fusion.}
Depth provides complementary geometric information to RGB images in a lot of tasks, such as semantic segmentation. Most state-of-the-art methods \cite{hu2019acnet, park2017rdfnet, seichter2020efficient, hazirbas2016fusenet, jiang2018rednet} for RGB-Depth semantic segmentation leveraged an RGB encoder and a depth encoder respectively, and then applied middle fusion between them to better incorporate low-level (texture, geometry) and high-level (semantic) features.
In the field of autonomous driving, there have also been great efforts on fusing camera frames and LiDAR scans for better object detection~\cite{mees2016choosing,Liang_2018_ECCV,prakash2021multi}.


\section{Conclusion}
In this paper, we identify a serious phenomenon called modality failure in multi-modal training. To tackle this issue, we propose Uni-Modal Teacher (UMT), which significantly improves the performances on both the audio-visual scene classification task and the RGB-Depth semantic segmentation task. Our experiments suggest several avenues for future works. For example, can we alleviate the modality failure by carefully designing the encoders, and whether a similar problem also exists in uni-modal learning? In addition, generalizing our method to other tasks including multi-modal object detection and generation will benefit many realistic applications.
We hope our findings will shed new light on multi-modal learning research.

\section{Broader impact}
\paragraph{Potential benefits.} Our UMT method improves multi-modal recognition and perception systems. In addition, our method enables multi-modal training with lower computational footprints, which can be a crucial step towards environment friendly AI. Finally, we provide theoretical analysis of our method, which tries to explain the current black-box multi-modal models, to facilitate transparency and equality in current AI research.







\bibliographystyle{plain}
\bibliography{refs}

\begin{thebibliography}{10}

\bibitem{afouras2018deep}
Triantafyllos Afouras, Joon~Son Chung, Andrew Senior, Oriol Vinyals, and Andrew
  Zisserman.
\newblock Deep audio-visual speech recognition.
\newblock {\em IEEE transactions on pattern analysis and machine intelligence},
  2018.

\bibitem{albanie2018emotion}
Samuel Albanie, Arsha Nagrani, Andrea Vedaldi, and Andrew Zisserman.
\newblock Emotion recognition in speech using cross-modal transfer in the wild.
\newblock In {\em Proceedings of the 26th ACM international conference on
  Multimedia}, pages 292--301, 2018.

\bibitem{allen2020towards}
Zeyuan Allen-Zhu and Yuanzhi Li.
\newblock Towards understanding ensemble, knowledge distillation and
  self-distillation in deep learning.
\newblock {\em arXiv preprint arXiv:2012.09816}, 2020.

\bibitem{arandjelovic2017look}
Relja Arandjelovic and Andrew Zisserman.
\newblock Look, listen and learn.
\newblock In {\em Proceedings of the IEEE International Conference on Computer
  Vision}, 2017.

\bibitem{atrey2010multimodal}
Pradeep~K Atrey, M~Anwar Hossain, Abdulmotaleb El~Saddik, and Mohan~S
  Kankanhalli.
\newblock Multimodal fusion for multimedia analysis: a survey.
\newblock {\em Multimedia systems}, 16(6):345--379, 2010.

\bibitem{aytar2016soundnet}
Yusuf Aytar, Carl Vondrick, and Antonio Torralba.
\newblock Soundnet: Learning sound representations from unlabeled video.
\newblock In {\em Advances in Neural Information Processing Systems}, 2016.

\bibitem{bucilu2006model}
Cristian Buciluǎ, Rich Caruana, and Alexandru Niculescu-Mizil.
\newblock Model compression.
\newblock In {\em Proceedings of the 12th ACM SIGKDD international conference
  on Knowledge discovery and data mining}, pages 535--541, 2006.

\bibitem{cartas2019does}
Alejandro Cartas, Jordi Luque, Petia Radeva, Carlos Segura, and Mariella
  Dimiccoli.
\newblock How much does audio matter to recognize egocentric object
  interactions?
\newblock In {\em Workshop of the conference of Computer Vision and Pattern
  Recognition (CVPR)}, 2019.

\bibitem{chen2020soundspaces}
Changan Chen, Unnat Jain, Carl Schissler, Sebastia Vicenc~Amengual Gari, Ziad
  Al-Halah, Vamsi~Krishna Ithapu, Philip Robinson, and Kristen Grauman.
\newblock Soundspaces: Audio-visual navigation in 3d environments.
\newblock In {\em Proceedings of the European Conference on Computer Vision
  (ECCV)}, 2020.

\bibitem{chen2021localizing}
Honglie Chen, Weidi Xie, Triantafyllos Afouras, Arsha Nagrani, Andrea Vedaldi,
  and Andrew Zisserman.
\newblock Localizing visual sounds the hard way.
\newblock In {\em Proceedings of the Conference on Computer Vision and Pattern
  Recognition (CVPR)}, 2021.

\bibitem{chen2020vggsound}
Honglie Chen, Weidi Xie, Andrea Vedaldi, and Andrew Zisserman.
\newblock Vggsound: A large-scale audio-visual dataset.
\newblock In {\em ICASSP 2020-2020 IEEE International Conference on Acoustics,
  Speech and Signal Processing (ICASSP)}, pages 721--725. IEEE, 2020.

\bibitem{cheuk2020nnaudio}
Kin~Wai Cheuk, Hans Anderson, Kat Agres, and Dorien Herremans.
\newblock nnaudio: An on-the-fly gpu audio to spectrogram conversion toolbox
  using 1d convolutional neural networks.
\newblock {\em IEEE Access}, 8:161981--162003, 2020.

\bibitem{ephrat2018looking}
Ariel Ephrat, Inbar Mosseri, Oran Lang, Tali Dekel, Kevin Wilson, Avinatan
  Hassidim, William~T Freeman, and Michael Rubinstein.
\newblock Looking to listen at the cocktail party: A speaker-independent
  audio-visual model for speech separation.
\newblock {\em ACM Transactions on Graphics (TOG)}, 37(4), 2016.

\bibitem{gan2020music}
Chuang Gan, Deng Huang, Hang Zhao, Joshua~B Tenenbaum, and Antonio Torralba.
\newblock Music gesture for visual sound separation.
\newblock In {\em Proceedings of the IEEE/CVF Conference on Computer Vision and
  Pattern Recognition}, pages 10478--10487, 2020.

\bibitem{gan2020look}
Chuang Gan, Yiwei Zhang, Jiajun Wu, Boqing Gong, and Joshua~B Tenenbaum.
\newblock Look, listen, and act: Towards audio-visual embodied navigation.
\newblock In {\em 2020 IEEE International Conference on Robotics and Automation
  (ICRA)}, pages 9701--9707. IEEE, 2020.

\bibitem{gao20192}
Ruohan Gao and Kristen Grauman.
\newblock 2.5 d visual sound.
\newblock In {\em Proceedings of the IEEE/CVF Conference on Computer Vision and
  Pattern Recognition}, pages 324--333, 2019.

\bibitem{gemmeke2017audio}
Jort~F Gemmeke, Daniel~PW Ellis, Dylan Freedman, Aren Jansen, Wade Lawrence,
  R~Channing Moore, Manoj Plakal, and Marvin Ritter.
\newblock Audio set: An ontology and human-labeled dataset for audio events.
\newblock In {\em 2017 IEEE International Conference on Acoustics, Speech and
  Signal Processing (ICASSP)}, pages 776--780. IEEE, 2017.

\bibitem{harwath2018jointly}
David Harwath, Adria Recasens, D{\'\i}dac Sur{\'\i}s, Galen Chuang, Antonio
  Torralba, and James Glass.
\newblock Jointly discovering visual objects and spoken words from raw sensory
  input.
\newblock In {\em Proceedings of the European conference on computer vision
  (ECCV)}, pages 649--665, 2018.

\bibitem{hazirbas2016fusenet}
Caner Hazirbas, Lingni Ma, Csaba Domokos, and Daniel Cremers.
\newblock Fusenet: Incorporating depth into semantic segmentation via
  fusion-based cnn architecture.
\newblock In {\em Asian conference on computer vision}, pages 213--228.
  Springer, 2016.

\bibitem{he2016deep}
Kaiming He, Xiangyu Zhang, Shaoqing Ren, and Jian Sun.
\newblock Deep residual learning for image recognition.
\newblock In {\em Proceedings of the IEEE conference on computer vision and
  pattern recognition}, pages 770--778, 2016.

\bibitem{hinton2015distilling}
Geoffrey Hinton, Oriol Vinyals, and Jeff Dean.
\newblock Distilling the knowledge in a neural network.
\newblock {\em arXiv preprint arXiv:1503.02531}, 2015.

\bibitem{hori2017attention}
Chiori Hori, Takaaki Hori, Teng-Yok Lee, Ziming Zhang, Bret Harsham, John~R
  Hershey, Tim~K Marks, and Kazuhiko Sumi.
\newblock Attention-based multimodal fusion for video description.
\newblock In {\em Proceedings of the IEEE international conference on computer
  vision}, pages 4193--4202, 2017.

\bibitem{hu2019acnet}
Xinxin Hu, Kailun Yang, Lei Fei, and Kaiwei Wang.
\newblock Acnet: Attention based network to exploit complementary features for
  rgbd semantic segmentation.
\newblock In {\em 2019 IEEE International Conference on Image Processing
  (ICIP)}, pages 1440--1444. IEEE, 2019.

\bibitem{huh2016makes}
Minyoung Huh, Pulkit Agrawal, and Alexei~A Efros.
\newblock What makes imagenet good for transfer learning?
\newblock {\em arXiv preprint arXiv:1608.08614}, 2016.

\bibitem{jiang2018rednet}
Jindong Jiang, Lunan Zheng, Fei Luo, and Zhijun Zhang.
\newblock Rednet: Residual encoder-decoder network for indoor rgb-d semantic
  segmentation.
\newblock {\em arXiv preprint arXiv:1806.01054}, 2018.

\bibitem{kingma2014adam}
Diederik~P Kingma and Jimmy Ba.
\newblock Adam: A method for stochastic optimization.
\newblock {\em arXiv preprint arXiv:1412.6980}, 2014.

\bibitem{korbar2018cooperative}
Bruno Korbar, Du~Tran, and Lorenzo Torresani.
\newblock Cooperative learning of audio and video models from self-supervised
  synchronization.
\newblock In {\em Proceedings of the Advances in Neural Information Processing
  Systems}, 2018.

\bibitem{Liang_2018_ECCV}
Ming Liang, Bin Yang, Shenlong Wang, and Raquel Urtasun.
\newblock Deep continuous fusion for multi-sensor 3d object detection.
\newblock In {\em Proceedings of the European Conference on Computer Vision
  (ECCV)}, September 2018.

\bibitem{mees2016choosing}
Oier Mees, Andreas Eitel, and Wolfram Burgard.
\newblock Choosing smartly: Adaptive multimodal fusion for object detection in
  changing environments.
\newblock In {\em 2016 IEEE/RSJ International Conference on Intelligent Robots
  and Systems (IROS)}, pages 151--156. IEEE, 2016.

\bibitem{morgado2018self}
Pedro Morgado, Nuno Vasconcelos, Timothy Langlois, and Oliver Wang.
\newblock Self-supervised generation of spatial audio for 360 video.
\newblock In {\em Advances in Neural Information Processing Systems}, 2018.

\bibitem{orsic2019defense}
Marin Orsic, Ivan Kreso, Petra Bevandic, and Sinisa Segvic.
\newblock In defense of pre-trained imagenet architectures for real-time
  semantic segmentation of road-driving images.
\newblock In {\em Proceedings of the IEEE/CVF Conference on Computer Vision and
  Pattern Recognition}, pages 12607--12616, 2019.

\bibitem{owens2018audio}
Andrew Owens and Alexei~A Efros.
\newblock Audio-visual scene analysis with self-supervised multisensory
  features.
\newblock In {\em Proceedings of the European Conference on Computer Vision},
  2018.

\bibitem{owens2016ambient}
Andrew Owens, Jiajun Wu, Josh~H. McDermott, William~T. Freeman, and Antonio
  Torralba.
\newblock Ambient sound provides supervision for visual learning.
\newblock In {\em Proceedings of the European conference on computer vision
  (ECCV)}, pages 801--816, 2016.

\bibitem{park2017rdfnet}
Seong-Jin Park, Ki-Sang Hong, and Seungyong Lee.
\newblock Rdfnet: Rgb-d multi-level residual feature fusion for indoor semantic
  segmentation.
\newblock In {\em Proceedings of the IEEE international conference on computer
  vision}, pages 4980--4989, 2017.

\bibitem{paszke2019pytorch}
Adam Paszke, Sam Gross, Francisco Massa, Adam Lerer, James Bradbury, Gregory
  Chanan, Trevor Killeen, Zeming Lin, Natalia Gimelshein, Luca Antiga, et~al.
\newblock Pytorch: An imperative style, high-performance deep learning library.
\newblock {\em arXiv preprint arXiv:1912.01703}, 2019.

\bibitem{perez2019mfas}
Juan-Manuel P{\'e}rez-R{\'u}a, Valentin Vielzeuf, St{\'e}phane Pateux, Moez
  Baccouche, and Fr{\'e}d{\'e}ric Jurie.
\newblock Mfas: Multimodal fusion architecture search.
\newblock In {\em Proceedings of the IEEE/CVF Conference on Computer Vision and
  Pattern Recognition}, pages 6966--6975, 2019.

\bibitem{prakash2021multi}
Aditya Prakash, Kashyap Chitta, and Andreas Geiger.
\newblock Multi-modal fusion transformer for end-to-end autonomous driving.
\newblock {\em arXiv preprint arXiv:2104.09224}, 2021.

\bibitem{romero2014fitnets}
Adriana Romero, Nicolas Ballas, Samira~Ebrahimi Kahou, Antoine Chassang, Carlo
  Gatta, and Yoshua Bengio.
\newblock Fitnets: Hints for thin deep nets.
\newblock {\em arXiv preprint arXiv:1412.6550}, 2014.

\bibitem{ronneberger2015u}
Olaf Ronneberger, Philipp Fischer, and Thomas Brox.
\newblock U-net: Convolutional networks for biomedical image segmentation.
\newblock In {\em International Conference on Medical image computing and
  computer-assisted intervention}, pages 234--241. Springer, 2015.

\bibitem{rouditchenko2019self}
Andrew Rouditchenko, Hang Zhao, Chuang Gan, Josh McDermott, and Antonio
  Torralba.
\newblock Self-supervised audio-visual co-segmentation.
\newblock In {\em ICASSP 2019-2019 IEEE International Conference on Acoustics,
  Speech and Signal Processing (ICASSP)}, pages 2357--2361. IEEE, 2019.

\bibitem{seichter2020efficient}
Daniel Seichter, Mona K{\"o}hler, Benjamin Lewandowski, Tim Wengefeld, and
  Horst-Michael Gross.
\newblock Efficient rgb-d semantic segmentation for indoor scene analysis.
\newblock {\em arXiv preprint arXiv:2011.06961}, 2020.

\bibitem{smith2005development}
Linda Smith and Michael Gasser.
\newblock The development of embodied cognition: Six lessons from babies.
\newblock {\em Artificial life}, 11(1-2):13--29, 2005.

\bibitem{srivastava2014dropout}
Nitish Srivastava, Geoffrey Hinton, Alex Krizhevsky, Ilya Sutskever, and Ruslan
  Salakhutdinov.
\newblock Dropout: a simple way to prevent neural networks from overfitting.
\newblock {\em The journal of machine learning research}, 15(1):1929--1958,
  2014.

\bibitem{wang2020makes}
Weiyao Wang, Du~Tran, and Matt Feiszli.
\newblock What makes training multi-modal classification networks hard?
\newblock In {\em Proceedings of the IEEE/CVF Conference on Computer Vision and
  Pattern Recognition}, pages 12695--12705, 2020.

\bibitem{wang2020deep}
Yikai Wang, Wenbing Huang, Fuchun Sun, Tingyang Xu, Yu~Rong, and Junzhou Huang.
\newblock Deep multimodal fusion by channel exchanging.
\newblock {\em Advances in Neural Information Processing Systems}, 33, 2020.

\bibitem{yang2020telling}
Karren Yang, Bryan Russell, and Justin Salamon.
\newblock Telling left from right: Learning spatial correspondence of sight and
  sound.
\newblock In {\em Proceedings of the IEEE/CVF Conference on Computer Vision and
  Pattern Recognition}, pages 9932--9941, 2020.

\bibitem{zhang2019your}
Linfeng Zhang, Jiebo Song, Anni Gao, Jingwei Chen, Chenglong Bao, and Kaisheng
  Ma.
\newblock Be your own teacher: Improve the performance of convolutional neural
  networks via self distillation.
\newblock In {\em Proceedings of the IEEE/CVF International Conference on
  Computer Vision}, pages 3713--3722, 2019.

\bibitem{zhao2018sound}
Hang Zhao, Chuang Gan, Andrew Rouditchenko, Carl Vondrick, Josh McDermott, and
  Antonio Torralba.
\newblock The sound of pixels.
\newblock In {\em Proceedings of the European conference on computer vision
  (ECCV)}, pages 570--586, 2018.

\bibitem{zhao2017pyramid}
Hengshuang Zhao, Jianping Shi, Xiaojuan Qi, Xiaogang Wang, and Jiaya Jia.
\newblock Pyramid scene parsing network.
\newblock In {\em Proceedings of the IEEE conference on computer vision and
  pattern recognition}, pages 2881--2890, 2017.

\end{thebibliography}






\end{document}